%% file: root.tex
\documentclass[letterpaper, 10 pt, conference]{ieeeconf}  % Comment this line out if you need a4paper

\IEEEoverridecommandlockouts                              % This command is only needed if 
\usepackage{caption}
\usepackage{subcaption}
\usepackage{graphicx}
\usepackage{amsmath}
\usepackage{amssymb}
\usepackage{algorithm}
\usepackage{algpseudocode}
\usepackage{xcolor}
\usepackage{bbm}
\newtheorem{theorem}{Theorem}
\newtheorem{cor}{Corollary}
\newtheorem{assump}{Assumption}

\renewcommand{\P}{\mathbb{P}}
\title{\LARGE \bf
Conformal Policy Learning for Sensorimotor Control\\ Under Distribution Shifts
}

\author{Huang Huang$^{1}$*, Satvik Sharma$^{1}$*, Antonio Loquercio$^{1}$*, \\Anastasios Angelopoulos$^{1}$, Ken Goldberg$^{1}$, Jitendra Malik$^{1}$
\thanks{*Equal contribution, $^{1}$ University of California, Berkeley}%
}

\begin{document}

\maketitle
\thispagestyle{empty}
\pagestyle{empty}

%%%%%%%%%%%%%%%%%%%%%%%%%%%%%%%%%%%%%%%%%%%%%%%%%%%%%%%%%%%%%%%%%%%%%%%%%%%%%%%%
\begin{abstract}

This paper focuses on the problem of detecting and reacting to changes in the distribution of a sensorimotor controller's observables.
The key idea is the design of switching policies that can take conformal quantiles as input, which we define as
\emph{conformal policy learning}, that allows robots to detect distribution shifts with formal statistical guarantees.
We show how to design such policies by using conformal quantiles to switch between base policies with different characteristics, e.g. safety or speed, or directly augmenting a policy observation with a quantile and training it with reinforcement learning.
% (the latter does not work as well as the former in our experiments, despite its conceptual appeal)
%Despite the latter being novel and an interesting form of learning under uncertainty, its practical performance is not as good as the switching policy, .
%
Theoretically, we show that such policies achieve the formal convergence guarantees in finite time.
In addition, we thoroughly evaluate their advantages and limitations on two compelling use cases: simulated autonomous driving and active perception with a physical quadruped.
Empirical results demonstrate that our approach outperforms five baselines.
It is also the simplest of the baseline strategies besides one ablation.
Being easy to use, flexible, and with formal guarantees, our work demonstrates how conformal prediction can be an effective tool for sensorimotor learning under uncertainty.

\end{abstract}

%%%%%%%%%%%%%%%%%%%%%%%%%%%%%%%%%%%%%%%%%%%%%%%%%%%%%%%%%%%%%%%%%%%%%%%%%%%%%%%%
\section{Introduction}
\label{sec:introduction}

% What is the problem?

\begin{figure}
    \centering
    \includegraphics[width=0.45\textwidth]{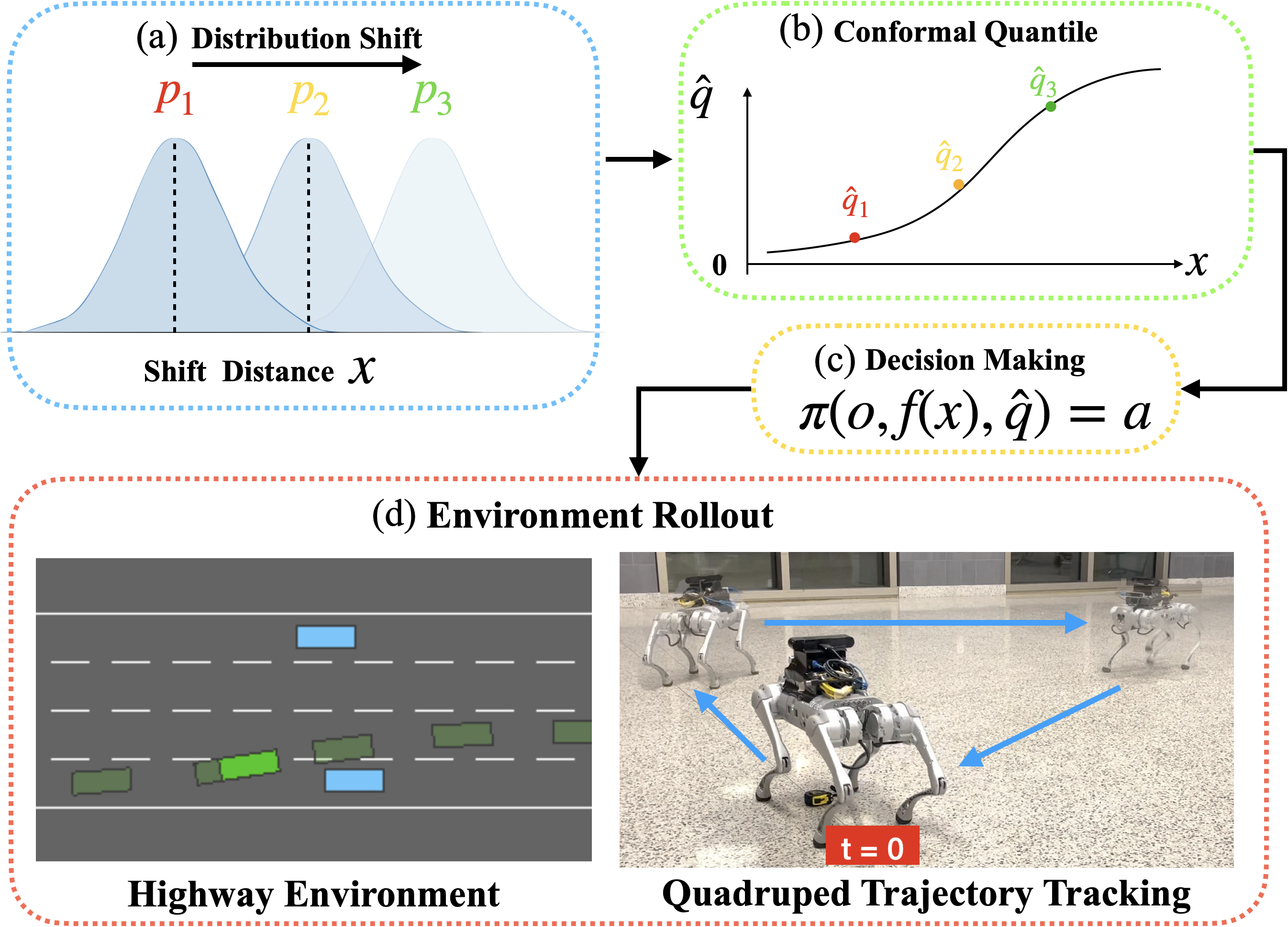}
    \caption{Illustration of conformal policy learning. As the distribution shifts (a), the uncertainty of the prediction increases, reflected by an increasing quantile (b). (c) Our policy takes into the observation $O$, prediction $f(x)$ and the quantile $\hat{q}$ to decide the actions to take. (d) We evaluate this framework on two use cases: autonomous driving in simulation and vision-based trajectory tracking with a physical quadruped.}
    \label{fig:fig1}
    \vspace{-10pt}
\end{figure}

As robots break out of lab-controlled conditions and start to operate for and around humans, there is a critical need to ensure their safety and reliability.
When operating in the wild, such robots increasingly rely on machine learning systems: in modular and end-to-end systems, there is a neural network at the core of the decision-making process interpreting high-dimensional observations, making predictions, or directly predicting actions.
However, when deployed in unstructured environments, such neural networks often encounter data distributions that change over time, negatively affecting their performance and, in turn, the safety of the overall system.
We present an approach that allows robots to quantify such distribution shifts with formal statistical guarantees, and to use this information downstream on control tasks (Fig.~\ref{fig:fig1}).

Traditional approaches for estimating distribution shifts model the network activations and weights by parametric probability distributions~\cite{hernandez2015probabilistic}, estimate uncertainties through sampling~\cite{Gal2015MCDO} or train an estimator of uncertainty by using a loss function~\cite{koenker1978regression, Kendall2017}.
However, these methods tend to generate over-confident predictions~\cite{osband2016risk} and provide neither rigorous statistical guarantees nor calibrated uncertainties.
Conformal prediction~\cite{vovk2005algorithmic} offers a principled approach to quantifying the uncertainty of black-box machine learning models, which has fueled a recent surge in its popularity~\cite{lei2018distribution, fisch2020efficient, romano2019conformalized, angelopoulos2020sets, romano2020classification, angelopoulos-gentle}.

Conformal prediction works as follows.
First, the user specifies a \emph{score function}, $s(x,y)$, that measures the quality of a prediction $f(x)$ from a black-box model.
For example, the score function can be the absolute distance between a prediction $f(x_i)$ and the ground truth $y_i$, i.e., $s(x_i, y_i) = | y_i - f(x_i)|$.
Then, given a sample of exchangeable calibration data points $\{(X_1, Y_1)\}_{i=1}^n$, and a new data point $(X,Y)$ with an unobserved label, a prediction set is formed for the label by inverting the score function as
\begin{equation}
    \mathcal{C}(X) = \{ y : s(X,y) \leq \hat{q} \},
\end{equation}
where $\hat{q}$ is chosen as the $(1-\alpha)(1+1/n)$-quantile of the scores on the calibration data.
This procedure results in prediction sets with valid coverage~\cite{vovk2005algorithmic, papadopoulos2002inductive}, 
\begin{equation}
    \P( Y \in \mathcal{C}(X) ) \geq 1-\alpha,
\end{equation}
where $\alpha$ is the error rate specified by the user.
Despite the restrictive assumptions that the data points are exchangeable~\cite{vovk2005algorithmic,papadopoulos2002inductive}, traditional conformal prediction has been applied in several robotics domains, e.g., to design sample-efficient alert systems for self-driving cars~\cite{luo2022sample} and object-pose estimation~\cite{yang2023object}, and language-based planning~\cite{ren2023robots}.
However, the data observed during the operation of a robot is more akin to a time series, whose distribution might change over time, due to various factors such as changes in the environment (night-day), sensor degradation, or unforeseen operating conditions.
Such conditions invalidate traditional conformal prediction since the data-generation process is not exchangeable.

Recently, new forms of conformal prediction have emerged to handle entirely non-exchangeable settings such as time-series prediction~\cite{gibbs2021adaptive,gibbs2022conformal, zaffran2022adaptive, bastani2022practical, bhatnagar2023improved, angelopoulos2023conformal}.
The standard setting, in this case, is the \emph{adversarial sequence model}, in which the data points $\{(x_t, y_t)\}_{t \in [T]}$ are arbitrary deterministic objects (e.g., real numbers) devoid of any probabilistic meaning.
Therefore, methods that can provide robustness in that setting are insulated against \emph{all possible realizations of future data} --- including those drawn by an omniscient adversary strategizing against the agent with full knowledge of its current and future plans.
Such adversarial settings have been popular since the early literature on calibration~\cite{dawid1982well} and were introduced into the realm of conformal prediction by~\cite{gibbs2021adaptive}.
Note that these approaches lose the guarantee at every time-step and instead provide \emph{long-term coverage}. That is, for some sequence of sets $\mathcal{C}_t$, letting $\text{err}_t = 1\{y_t \notin \mathcal{C}_t(x_t) \}$, they achieve
\[
\frac{1}{T} \sum_{i=1}^{T} \text{err}_t = \alpha + o(1).
\].

Such a setting, more akin to real-world conditions, has quickly captured the attention of roboticists for applications in multi-agent motion-planning problems~\cite{dixit2023adaptive, lindemann2023safe, muthali2023multi}.
In these examples, prediction sets for the other agents' positions, are incorporated as safety constraints into the planner.
However, simpler conformal methods with tighter prediction sets and greater stability have been developed---namely, the conformal PID control method~\cite{angelopoulos2023conformal}. 
This approach for generating prediction sets has only been applied in traditional time-series prediction settings, where the data distribution at time $t$ is unaffected by the predictions and decisions made at times $1, \ldots, t-1$. 
Furthermore, the utility of these prediction sets are somewhat limited, as they require a robotic system in which constraints are easy to incorporate.
Ad exemplum, policies trained with deep learning \emph{cannot} accept the sort of mathematical constraints that are common in optimization.

The main methodological innovation of this paper is the design of policies that can take conformal quantiles as input; we call this \emph{conformal policy learning}.
Our results indicate it is possible to train reinforcement learning policies that, in practice, satisfy the conditions needed~\cite{lekeufack2023conformal} to achieve formal coverage guarantees in finite time; we call such policies \emph{conformal policies} and show the first practical examples of their existence.
%Distribution-free guarantees are available also via a \emph{switching policy} which can be run separately or in parallel.
%On the practical side, we provide validation of this methodology in decision-making problems under distribution shifts, giving evidence of its utility for robotics.
%This is the first physical experiment testing conformal controllers and conformal decision theory.

To give a bit more detail, we explore conformal policies of two forms.
The first is a naive switching mechanism, falling back to a safe policy whenever the quantile exceeds a user-defined threshold~\cite{loquercio2020general}.
The switching policy has safety guarantees under the relatively weak assumption that the safe policy is \emph{eventually safe}, i.e., that it will have low risk when deployed for long enough~\cite{lekeufack2023conformal}.
However, its reliance on a hand-designed threshold for switching between policies can make tuning challenging.
In the second, we aim to remove such user-defined threshold by directly augmenting the observation of a policy with conformal quantiles, and training such policy with reinforcement learning.
To our knowledge, this strategy is novel and demonstrates an interesting new form of learning under uncertainty, although its practical performance is not yet as good as that of the switching policy.

We demonstrate these ideas in two compelling use cases: the first is a simulated autonomous driving scenario, where an autonomous car must drive as fast as possible while avoiding other vehicles.
The second is a real-world active perception problem with a quadruped, where the goal is to minimize a trajectory tracking error by actively controlling the robot's speed over the trajectory to maximize the quality of vision-based state estimation.
We compare against many strong baselines in simulation and show that our approach empirically achieves the desired guarantees while significantly outperforming these baselines.
In addition, we report the interesting finding that learning a policy with conformal quantiles as input systematically underperforms the simple switching strategy in out-of-distribution settings, indicating that the learning procedure fails to capture the relation between the quantile and the overall policy uncertainty.
We provide an in-depth study of this phenomenon, a set of hypotheses underpinning this behavior, and possible avenues to improve its performance.

\section{Related Work}

Robotics research has allocated a lot of attention to decision-making under uncertainty in the context of machine learning approaches~\cite{sunderhauf2018limits}.
Given the recent surge in popularity of conformal prediction, several works showed how to use it to add statistical guarantees in perception~\cite{yang2023object} and planning~\cite{dixit2023adaptive, lindemann2023safe, muthali2023multi,luo2022sample}.
Recent work shows that conformal quantiles can control the hallucination of large language models, making the latter better in learning-based planning~\cite{ren2023robots}.
However, these approaches do not provide exhaustive studies on how conformal quantiles could be used directly for control in real-world robotics scenarios.

Conversely, PAC-Bayes generalization theory has been used for deriving conditional bounds to influence decision-making on vision-based aerial navigation and manipulation in Farid et al~\cite {farid2022failure}.
However, this approach focused on studying generalization \emph{in the training distribution}, while our work focuses on studying the problem of generalization under distribution shifts.
In addition, they lack a study on how such bounds could be directly used by a policy trained with reinforcement learning.

In robotics, uncertainty has not only been used in the context of safety but also to speed up the reinforcement learning on a real robot~\cite{kahn2018self}, combine optimal control with neural networks~\cite{kaufmann2018beauty, loquercio2020general}, or to increase the efficiency of learning in manipulation~\cite{chua2018deep}.
However, these works neither provide formal guarantees nor an easy and interpretable way to change the behavior of the control policy as a function of its safety.
Conversely, our work can use conformal quantiles to directly affect the controller's performance with a single and interpretable user-specified parameter, $\alpha$, i.e., the desired error rate. This gives the user the possibility to have more greedy behavior, hence favoring exploration, or a more conservative attitude, promoting safety.
In addition, we empirically show to achieve the desired error rate, as predicted by the underlying theory.

\section{Method}
We describe our methodology, first focusing on its theoretical foundation in conformal PID control and then develop the idea of conformal policy learning --- i.e., the design of policies that take conformal quantiles as input.

\subsection{Conformal control methodology}
\label{sec:conformal_control}

\textbf{Producing the sets.} Similarly to classical conformal prediction, we will proceed at time $t$ by forming prediction sets $\mathcal{C}_t$ in the following way:
\begin{equation}
    \label{eq:set-construction}
    \mathcal{C}_t = \{ y : s(x_t, y) \leq q_t \},
\end{equation}
where the quantile $q_t$ is up to us to choose.
The set construction in~\eqref{eq:set-construction} is almost exactly the same as that of standard conformal prediction; the difference is that $q_t$ is \emph{not} an empirical quantile of the previous data points.
Instead, it will be dynamically modulated using conformal P control, i.e., quantile tracking~\cite{angelopoulos2023conformal}.
\textit{Why is the parameter $q_t$ so important?} 
Because the miscoverage event $\text{err}_{t}$ abides by the equivalence $\text{err}_t = 1 \Longleftrightarrow s(x_t, y_t) > q_t$.
Therefore, picking the right sequence of quantile updates is critical to achieving coverage.

\textbf{Tracking the quantile.} The update for $q_t$ takes the following form:
\begin{equation}
    \label{eq:quantile-update}
    q_{t+1} = q_t - \eta \nabla \rho_{1-\alpha}(\text{err}_t - \alpha),
\end{equation}
where $\rho$ is the quantile loss (sometimes referred to as the ``pinball loss'') at level $1-\alpha$.
This update can be seen as a simple form of P-control where the feedback signal is $\text{err}_t$ and we would like it close to $1-\alpha$.

\textbf{Proof of validity.} A simplified and slightly improved proof of the result in~\cite{angelopoulos2023conformal} is given (this analysis strategy was first developed in~\cite{gibbs2021adaptive}).
\begin{theorem}
    Let the scores $s(x_1, y_1), \ldots, s(x_T, y_T)$ be bounded between $[0,B]$.
    Then we have, uniformly for all positive integers $W$ and $T_0$ satisfying $T_0 + W \leq T$ and all realizations $(x_1, y_1), \ldots, (x_T, y_T)$, that
    \begin{equation}
        \frac{1}{W} \sum\limits_{t=T_0}^{T_0 + W} \text{err}_t \leq \alpha + \frac{B + \eta}{\eta W}.
    \end{equation}
\end{theorem}
% \begin{proof}
\proof
    Akin to Proposition 5 of~\cite{gibbs2021adaptive}, we telescope the sum in~\eqref{eq:quantile-update} to get 
    \begin{equation}
        \begin{aligned}
            \frac{q_{T_0 + W} - q_{T_0}}{\eta W}  &= \frac{1}{W} \sum\limits_{t=T_0}^{T_0 + W}\nabla \rho_{1-\alpha}(\text{err}_t - \alpha) \\ &= \frac{1}{W} \sum\limits_{t=T_0}^{T_0 + W} \text{err}_t - \alpha,
        \end{aligned}
    \end{equation}
    where the last step used the definition of the gradient of the quantile loss. 
    Rearranging terms, the implication is that
    \begin{equation}
        \frac{1}{W} \sum\limits_{t=T_0}^{T_0 + W} \text{err}_t \leq \alpha  + \frac{q_{T_0 + W} - q_{T_0}}{\eta W} \leq \alpha + \frac{B+\eta}{\eta W},
    \end{equation}
    where the last inequality holds because $q_{T_0 + W} \leq B + \alpha \eta$ and $q_{T_0} \geq -\eta(1-\alpha) $ .
% \end{proof}
\endproof
    The nature of the result is~\emph{multi-resolution} in the sense that for all possible windows in time, the coverage will be met, and has a finite sample correction at level $(B+\eta)/\eta W$.
    In practice, we use a (slightly) adaptive version of the quantile tracker with $\eta_t = 0.1\hat{B}_t$.
    It is trivial to get a guarantee in this setting as well.
    Furthermore, a two-sided guarantee is available by a similar proof strategy.

\subsection{Conformal Policies for Sensorimotor Control}
\label{sec:conformal_control_policy}

We introduce the formal framework of a conformal policy.
The policy is a sensorimotor control algorithm $\pi(x_s, f, q)$ that observes a state $x_s$, a conformal quantile $q$, and a predictor $f$.
The latter predicts future values that could influence decision-making, e.g. the future geometry of the terrain for locomotion~\cite{loquercio2023learning}, the future position of other agents~\cite{bajcsy2023learning}, or the low-level dynamics of the robot~\cite{smith2023grow}.
We do not make any specific assumption on the nature of the policy or the predictor.
For a conformal policy to have formal safety guarantees, it must be \emph{eventually safe}:
\begin{assump}[Eventually safe.]
    \label{assn:eventually-safe}
    A policy $\pi(x_s, f, q)$ is eventually safe if there exist $\alpha^{\rm safe} < \alpha$, $q^{\rm safe}$, and $K \in \mathbb{N}$ satisfying
    \begin{equation}
        \begin{aligned}
        \big\{\forall k\in [K], q_k & \geq q^{\rm safe}\big\} \\
        & \Longrightarrow \frac{1}{K} \sum_{k=1}^K \mathbf{1}\{s(x_k,y_k) > q_k\} \leq \alpha^{\rm safe}.% \\
        \end{aligned}
    \end{equation}
    Here, the policy is run with quantile $q_k$ to produce the values $(x_{k+1}, y_{k+1})$.
\end{assump}
The assumption may at first look strange because the inequality does not involve $\pi$. 
However, one must remember that the actions of $\pi$ produce the future values of $x_k$ and $y_k$.
The assumption therefore encodes the idea that if the policy makes enough errors, it can revert to a situation where its prediction error is lower.
For example, the robot can slow down, making the task of perception and scene understanding easier.
Under this assumption, the sets will cover at the correct frequency:
\begin{cor}
    Under Assumption~\ref{assn:eventually-safe}, we have that 
    \begin{equation}
        \frac{1}{T} \sum\limits_{t=1}^T \text{err}_t \leq \alpha + o(1).
    \end{equation}
\end{cor}
% \begin{proof}
\proof
    Assumption~\ref{assn:eventually-safe} implies the decision policy is safe over time~\cite{lekeufack2023conformal} with respect to the miscoverage loss.
    By Theorem 1 of the same, the miscoverage loss is controlled.
% \end{proof}
\endproof
The validity of switching policies is also addressed as the special case of this proposition.

Of course, it is worth noting that a tacit assumption in our problem setup is the existence of online feedback, i.e., that the predictor can verify its forecasts against the ground-truth after a limited amount of time.
Though not entirely benign, this assumption is valid in several real-world scenarios.
For example, a legged robot will observe whether its predictions about the terrain geometry are valid after walking over it~\cite{loquercio2023learning}, or an agent could verify the accuracy of trajectory predictions for other agents several time steps after the prediction intervals are formed~\cite{bajcsy2023learning}.
% 

%We use conformal tracking (Sec.~\ref{sec:conformal_control}) to estimate the $1-\alpha$ quantile $q_t$ of $f(x_o)$. We use such information in two ways: (1) training a conformal policy $\pi_c(x_s,f(x_o),q_t)$ with reinforcement learning, and (2) switching between predefined behaviors, i.e.

% estimated by the previous algorithm downstream in two different ways. The first is a switching mechanism based on the :
% \[
%     f(x)= 
% \begin{cases}
%     \frac{x^2-x}{x},& \text{if } x\geq 1\\
%     0,              & \text{otherwise}
% \end{cases}
% \]

\section{Simulation Experiments}

\subsection{Experimental Setup}
We evaluate our approach in a simulation highway environment, where an autonomous driving agent must drive as fast as possible while avoiding other vehicles~\cite{highway-env}.
We instantiate the environment with 4 lanes, 50 other cars, and the ego vehicle. The action space of the ego-vehicle consists of 5 discrete actions: switching to the left lane, switching to the right lane, speeding up, slowing down, and maintaining the current speed. At each time step, the observation consists of the relative x and y positions and relative x and y velocities of the closest five vehicles to the ego-vehicle, normalized between -1 and 1. The ego-vehicle can only observe other agents in front of it. A visualization of the environment is shown in Figure \ref{fig:fig1}. The agent can move in a specified speed range. A rollout stops when the ego-vehicle collides with other vehicles or after 40 seconds.
% The ego-vehicle can drive at the speed in a specified range. Each rollout terminates when the ego-vehicle collides with the surrounding vehicles or after 40s.
\subsection{Base Policies Training}
%\textbf{Base Policies Training.}
We consider two base policies. A policy trained only with a non-collision reward, denoted as $\pi_{safe}$, and a policy trained with both the non-collision reward and the speed reward, $\pi_{speed}$. The minimum speed of the ego-vehicle is 20$m/s$, so no policy can immediately stop.
%the objective of $\pi_{safe}$ is to avoid collision with other vehicles while keep driving on the highway, regardless of the speed. 
%Therefore $\pi_{safe}$ will have lower chance of colliding compared to $\pi_{speed}$, which would run faster but have higher chance of colliding.
We train both $\pi_{safe}$ and $\pi_{speed}$ with Deep Q-Network (DQN)~\cite{mnih2015human} for 200K steps. For $\pi_{safe}$, we have a collision reward of -1 and for $\pi_{speed}$, we have an additional high-speed reward of 0.5. Both policies are 2-layer MLPs with 256 neurons for each layer. %We use these two policies as the primitives in our experiments.

\subsection{Predictor Training}
\label{ssub:pred_train}
We train a predictor to forecast the nearest distance to the surrounding vehicles $l$ three steps into the future. The input to the predictor is the history of the past five observations. We normalize $l$ into the $[0,1]$ interval by applying the following transformation $\Tilde{l} = 2/(1+e^{(0.1l)})$.
%
%Specifically, at each time step $t$, the predictor takes an hist and predicts $\Tilde{l}$ in $t+3$, where $\Tilde{l} = \frac{2}{1+e^(0.1*l)}$. As $l\geq 0$, $\Tilde{l}$, which is the sigmoid of $l$, is always between 0 and 1. 
We collect a dataset of 30k rollouts with $\pi_{speed}$ and $\pi_{safe}$. During the data collection, the ego-vehicle speed range is between 20$m/s$ and 30$m/s$. We use a 3-layer MLP followed by a Sigmoid layer as the predictor with 64, 128 and 64 neurons for each layer. The predictor is trained until convergence with an MSE loss and a learning rate of 0.001. 

\subsection{Conformal Policies}
%Using the prediction $\Tilde{l}$ from the predictor and the quantile $\hat{q}$ from the conformal PID controller,
We use two policies for evaluation.
The first, $\pi_{switch}$, switches between $\pi_{safe}$ and $\pi_{speed}$ as a function of a threshold $q_{\rm safe}$ of the predicted future distance $\Tilde{l}_t$ and its estimated quantile $\Tilde{q}_t$ at $1-\alpha = 90\%$ coverage computed with~\cite{angelopoulos2023conformal}. Specifically,
\begin{equation}
    \pi_{switch}(x,\Tilde{l}_t,\Tilde{q}_t) = 
\begin{cases}
    \pi_{safe}(x),& \text{if } (\Tilde{l}_t + \Tilde{q}_t)\geq q_{\rm safe} \\
   \pi_{speed}(x),              & \text{otherwise.}
   \label{eq:switch}
\end{cases}
\end{equation}
Note that this policy satisfies Assumption~\ref{assn:eventually-safe} by design.
The second policy, $\pi_{RL}(x,\Tilde{l}_t,\Tilde{q}_t)$, is directly trained with DQN with the same reward of $\pi_{speed}$.
Ideally, $\pi_{RL}$ will automatically learn to modulate its speed as a function of its uncertainty. Note, however, that this design is not strictly guaranteed to satisfy Assumption~\ref{assn:eventually-safe}.

% a policy $\pi$ using $\pi_{safe}$ and $\pi_{speed}$ as the primitives to achieve a trade-off between safety and speed. At each time step $t$, $\pi$ takes in $\Tilde{l}$ and $\hat{q_t}$. If $\Tilde{l}+\hat{q_t} \geq \gamma$, where $\gamma=0.3$ is a user specified threshold, $\pi$ will execute the action given by $\pi_{safe}$. Otherwise, $\pi$ will execute the action given by $\pi_{speed}$. Intuitively, when $\Tilde{l}$ is large, the nearest distance between the ego-vehicle and the surrounding vehicles is small, indicating a high likelihood of colliding. $\pi_{safe}$ therefore should be executed to ensure safety. $\hat{q_t}$ provides additional uncertainty information.

\subsection{Baselines}
We compare against the following baselines for uncertainty estimation of $\Tilde{l}$.
\textbf{Bayesian Dropout~\cite{Gal2015MCDO}:} We add a 0.1 dropout rate at each linear layer of the predictor and compute uncertainties from five runs. \textbf{Ensemble~\cite{lakshminarayanan2017simple}:} We train five networks with five disjoints subsets of the data. \textbf{HAU~\cite{Kendall2017}:} Heteroscedastic Aleatoric Uncertainty (HAU). \textbf{No Quantile}: This is equivalent to $\pi_{switch}$ but without quantiles in~\eqref{eq:switch}. Essentialy, the policy switches to $\pi_{safe}$ if $\Tilde{l}_t\geq q_{safe}$.
\textbf{Orthonormal Certificates~\cite{taga2019}:} We learn a set of linear functions that map the training data to zero. We use the predictor's second-to-last hidden layer as a feature representation for the input. We project these features to a 100-dimensional space representing the linear certificates. We train this additional linear layer with the target label being 0 for 100 epochs.

\begin{figure}
    \centering
    \includegraphics[width=0.45\textwidth]{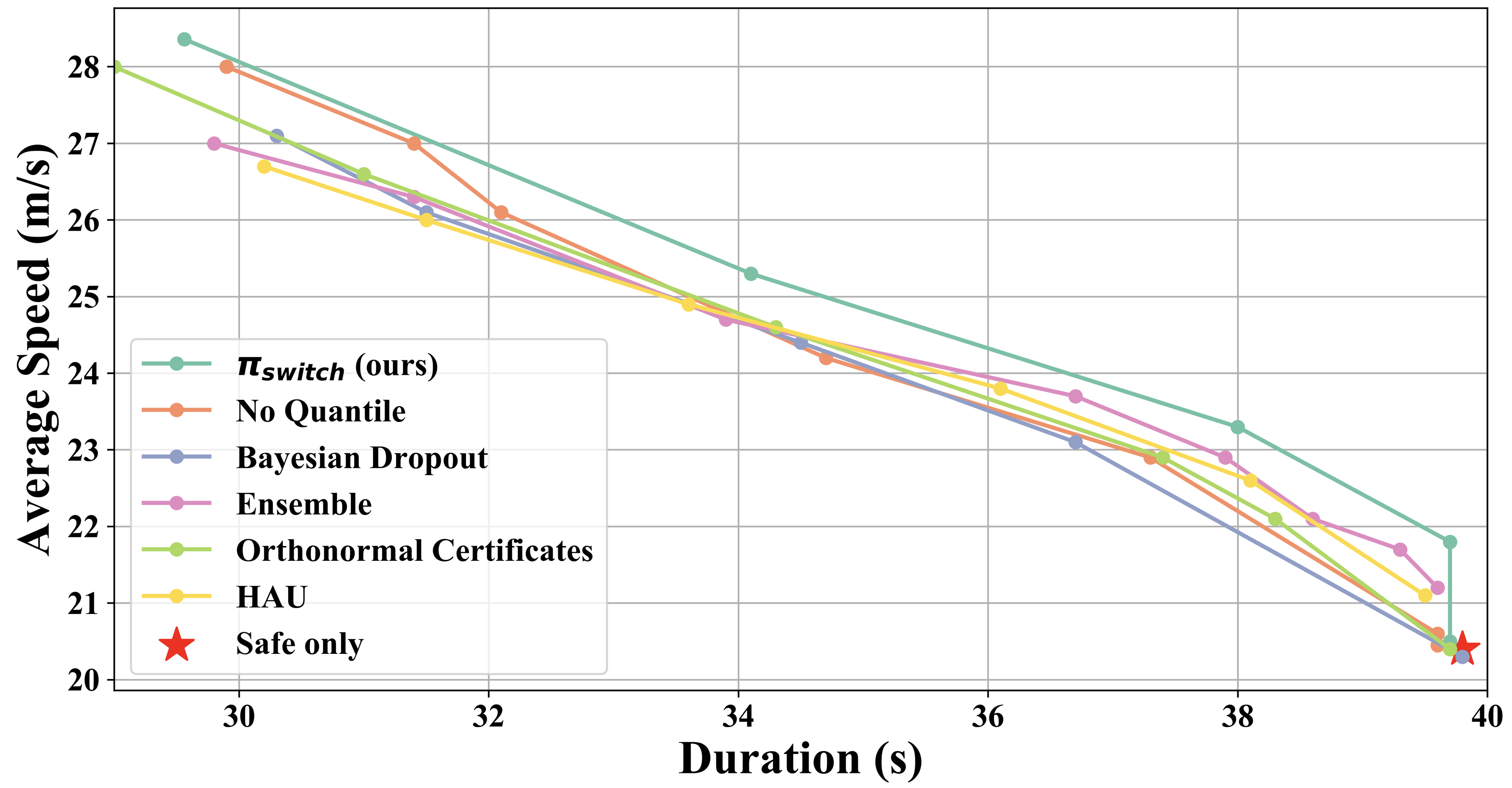}
    \caption{Pareto plot for the simulation environments comparing the performance of $\pi_{switch}$ to other baselines. Each method has its own Pareto curve. Each Pareto curve is constructed by varying the switching threshold $q_{safe}$. Each point in the curve is the average speed and the episode duration (an episode terminates if there is a collision or reaches 40 seconds) at a particular $q_{safe}$ over 500 realizations. The red star is the performance of $\pi_{safe}$. Our approach always achieves a higher speed than the baselines at the same collision frequency.}
    \label{fig:pareto}
    \vspace{-20pt}
\end{figure}

\subsection{Evaluation Results}
\begin{figure}
    \centering
    \includegraphics[width=0.45\textwidth]{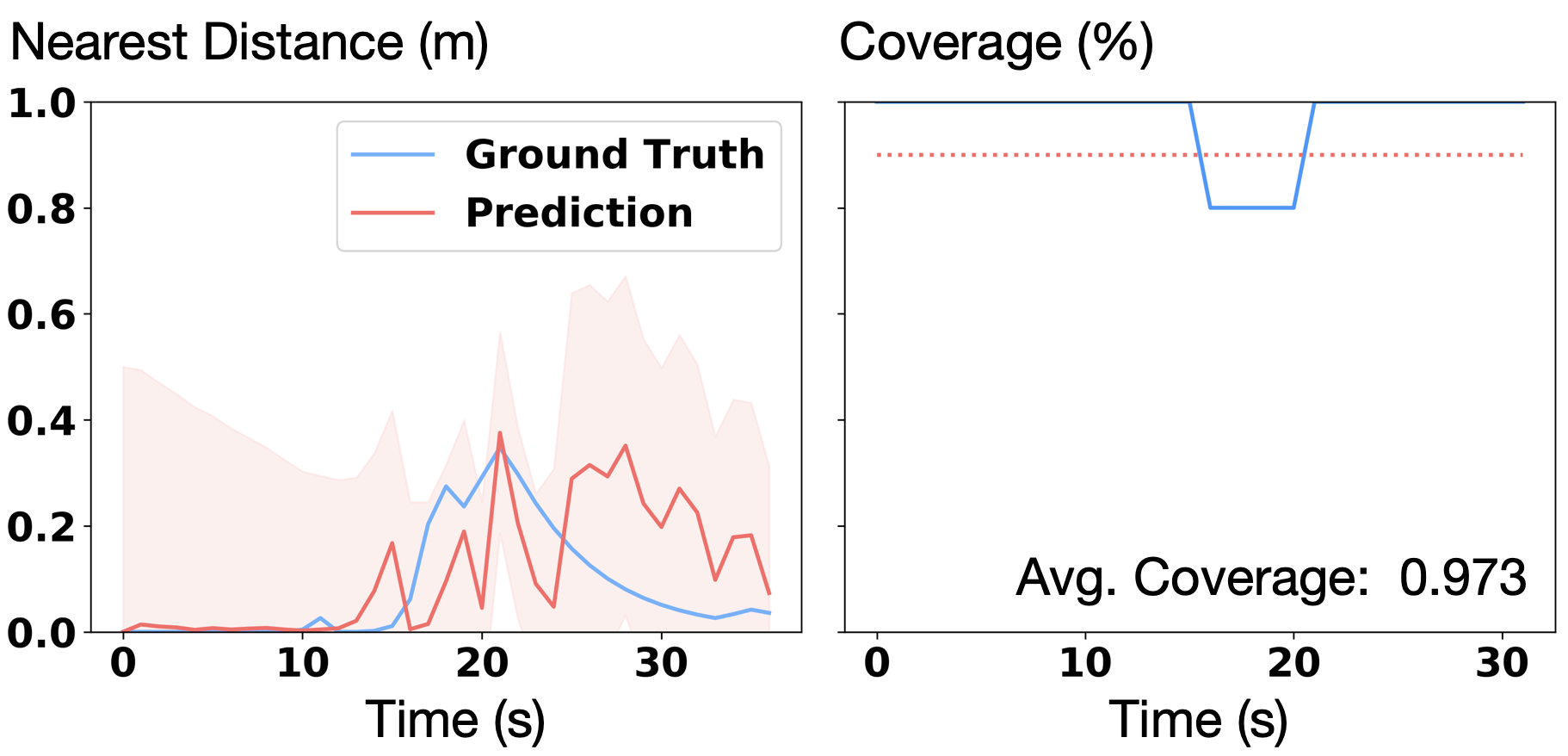}
    \caption{\textbf{Left}: Prediction and conformal quantile estimation in one simulation rollout for speed range of [20,30]$m/s$. The ground truth value of $\Tilde{l}$ is shown in blue and the prediction and quantile are shown in red. \textbf{Right}: Coverage with a sliding window size of 5 time steps. We show the coverage for $\Tilde{l}$ for speed range of [20,30]$m/s$, which is higher than the specified value 0.9 (red dotted line) most of the time.}
    \label{fig:coverage}
    \vspace{-20pt}
\end{figure}

 We consider two metrics: the average rollout duration and the ego-vehicle's average rollout speed. We compute the metrics by averaging the results over 500 different environment realizations.
 %which is the time of the episode before the ego-vehicle potentially collides with another vehicle, 
 %and the speed, which is the average speed of the ego-vehicle.
 We train the predictor and the base policies within a speed range of $[20,30]m/s$.
 We then evaluate all baselines by changing the speed range to $[20,40]m/s$. This will result in the policy observing both in and out of distribution states during evaluation.
 % At the evaluation time, we consider that the ego-vehicle speed range from 20$m/s$ to 30$m/s$ to be in distribution and  We evaluate two methods for decision making or $\pi(.)$ in Figure \ref{fig:method}. In Table \ref{table:baselines}, we evaluate the heuristic policy which switches between safe and speed policies depending on a threshold on the quantile. The results in Table \ref{table:baselines} are over 500 rollouts and report the average and standard deviations for the duration, speed, and distance for the conformal heuristic controller, the heuristic controller that does not utilize the quantile and only the predictor, and various baselines. 
 At evaluation time, we vary the switching threshold $q_{\rm safe}$ in Eq.~\eqref{eq:switch} to study the trade-off between safety and speed that each method provides. The results of these experiments are reported in Fig.~\ref{fig:pareto}. Our approach consistently outperforms the baselines and enables faster driving for each safety constraint. 
 The improvement provided by our approach is larger for more stringent values of $q_{safe}$, indicating that it better detects and reacts to dangerous conditions. Finally, our approach adds relatively no extra computation (only the update of Eq.~\ref{eq:quantile-update}) and requires no changes to the model, and yet outperforms ensemble methods, which increase the runtime computation budget fivefold. Beyond better performance, our approach also provides a formal guarantee of $1-\alpha=90\%$ coverage, which we verify empirically in Fig.~\ref{fig:coverage}.
 %it is interesting that our approach outperforms all the ensemble methods w

In Table~\ref{table:policy}, we compare different variants of conformal policies trained end-to-end with RL to $\pi_{speed}$ and $\pi_{switch}$. In the following, we describe the details of these baselines.
\textbf{$\pi_{RL}$} represents an agent that observes, in addition to the state $x_t$, the future prediction $\Tilde{l}_t$ and its conformal quantile $\Tilde{q}_t$.
The most important baseline for $\pi_{RL}$ is the agent \textbf{$\pi_{RL}^{nq}$}, which observes only $\Tilde{l}_t$ and has no access to quantile uncertainties.
We additionally ablate a set of reward designs for $\pi_{RL}$ (all the following have access to both $\Tilde{l}_t$ and $\Tilde{q}_t$).
\textbf{$\pi^1_{RL}$} is trained with an additional penalty on the norm of $\Tilde{q}_t$. This should push the agent to stay in known states, actively avoiding distribution shifts.
The agent \textbf{$\pi^2_{RL}$} is trained with a penalty of $\Tilde{q}_t \cdot \Tilde{l}_t$. This should push the agent only to become conservative when the probability of collision rises (due to an increase in $\Tilde{l}_t$), and quantile regression detects the state as out of distribution. The rationale is that, in distribution, the agent does not need to minimize the distance to other cars.
\textbf{$\pi^3_{RL}$} is trained with both a penalty on the norm $\Tilde{q}_t$ and a penalty on $\Tilde{q}_t \cdot \Tilde{l}_t$.
For the latter three agents, we keep the basic reward constant but select, for each agent separately, the weighting coefficient of the penalty terms to maximize their performance.

\input{table}

All policies are trained in the speed range $[20,30]m/s$ with DQN for 200K iterations and evaluated in three different speed ranges, increasingly out of the training distribution.
Table~\ref{table:policy} indicates that $\pi_{switch}$ is the safest agent by a significant margin.
The second-best performer is $\pi_{RL}$, with a 25\% longer survival time in the $[20,40]m/s$ range than the other ablations.
Nonetheless, no agents trained with RL manage to use the predictions and quantiles as effectively as a switching policy. This indicates that traditional RL algorithms can fail to account for the \emph{known unknowns}. Our hypothesis for this behavior is that at training time (by definition), the agent only sees data in distribution, thereby with a small $\Tilde{q}_t$. Therefore, it does not learn how to act when $\Tilde{q}_t$ is large.
Understanding how to make the policy take further advantage of the extra information and outperform our hand-designed $\pi_{switch}$ is an interesting avenue for future work.

%show that all variants of $\pi_{RL}$ fail to capture the correlation between the predictor's uncertainty and the quantiles. $\pi_{RL}$ performs similarly to $\pi_{RL}^{nq}$ and $\pi_{speed}$, for speed range of [20,30]$m/s$ and [20,50]$m/s$ and is slightly better for speed range of [20,40]$m/s$. Directly augmenting observations works better than modifying rewards with $\hat{q}$. $\pi_{switch}$ outperforms all other methods and can keep the desired level of safety in out-of-distribution data. We hypothesize that policies trained with RL learn the data distribution during training and estimate predictions and uncertainties internally. Therefore, there is no direct incentive to use explicit uncertainties. Promoting the correlation more effectively in policy learning is an interesting venue for future work.

% we directly input the observation, the prediction, and the quantile to a learned policy trained with DQN for 600 K epochs and compare this to a policy that only has access to the observation and the prediction.
% We can see from Table \ref{table:policy} that the conformal heuristic controller 

\section{Physical Experiments}
\input{slam_fig}
\subsection{Experimental Setup}
We evaluate our approach on an active perception problem on a physical quadruped. In these experiments, we only consider conformal policies with switching behaviors, i.e., $\pi_{switch}$, given their better performance with respect to end-to-end conformal policies like $\pi_{RL}$, as shown in the previous simulation experiments.

We attach a RealSense T265 tracking camera on the front of a Unitree Go1 quadruped. We use the built-in SLAM module of the camera to estimate the quadruped state. The objective is to have the quadruped follow a given trajectory as accurately and fast as possible. As the robot moves fast, the built-in SLAM module may generate SLAM errors, meaning the algorithm cannot estimate the current robot state. Such errors are correlated to the quadruped planned path and speed. We design $\pi_{switch}$ to speed up or slow down the quadruped to trade off the trajectory tracking quality with speed.

We use the built-in MPC to control the quadruped by providing the robot linear velocity commands in the x and y direction and the yaw angular velocity command. Given a tracking trajectory consisting of waypoints $(x_i,y_i,\theta_i), i\in [0,..., N]$, where $x_i,y_i, \theta_i$ is the robot coordinates and orientation in the world frame, the robot commands are computed by a P controller such that $v_x(t) = kp_x\cdot (\hat{x_t}-x_i),v_y(t) = kp_y\cdot (\hat{y_t}-y_i),v_\theta(t) = kp_{\theta}\cdot (\hat{\theta_t}-\theta_i)$. $(\hat{x_t},\hat{y_t},\hat{\theta_t})$ are the estimated robot coordinates and orientations given by SLAM, and $(x_i,y_i,\theta_i)$ is the nearest waypoint from the given trajectory to the current robot state. By adaptively changing P gains $kp_x, kp_y, kp_\theta $, our policy can speed up or slow down the quadruped to track fast or avoid SLAM errors, respectively. The policy runs at 10 Hz.

\subsection{Predictor Training}
We train a predictor to forecast whether the SLAM system will issue a tracking error in the future. Specifically, the predictor takes in the observation of 
% $(\hat{x_{t,..t-4}},\hat{y_t},\hat{\theta_t},x_{i,..,i+5},y_{i,..,i+5},\theta_{i,..,i+5}, v_x(t),v_y(t), v_\theta(t), )$
robot states and velocity commands of the past 5 steps, the current nearest waypoint, the next 3 waypoints to track, and SLAM errors received in the past 5 steps. Then, it predicts whether a SLAM error will occur at time step $t+2$, which we define as $\Tilde{l}_t$. The predictor is trained on a dataset collected on 43 randomly sampled trajectories. Each trajectory is generated by linearly interpolating two keypoints uniformly sampled from $[-2,2]m$ to create a zig-zag trajectory. The robot follows the trajectory with a constant $kp$ of 1.5. The predictor is a one-layer MLP with 128 neurons and a single output. As there are more negative than positive samples, we train the predictor with weighted binary cross entropy: the positive samples' loss is multiplied by a factor 5.

\subsection{Policy Design}
Our policy $\pi_{switch}$ takes the prediction $\Tilde{l}_t$  and the quantile $\hat{q}_t$ at $1-\alpha=80\%$ coverage given by the conformal PID controller.  $\pi_{switch}$  uses these two to modulate the robot's speed.
Specifically, when $\Tilde{l}_t+\Tilde{q}_t \geq q_{safe}$, $\pi_{switch}$ slows down the quadruped by updating $kp(t) = clip(0.8* kp(t-1), kp_{min}, kp_{max})$. Conversely, when $\Tilde{l}_t+\Tilde{q}_t \leq q_{safe}$, i.e. the probability of failure is low, the policy speeds up the robot by increasing the P gain to $kp(t) = clip(1.1* kp(t-1), kp_{min}, kp_{max})$. $q_{safe}=0.8$ is a specified threshold and $kp_{min}, kp_{max}$ are the lower and upper bound of $kp$. In practice, we notice the predictor can give many false positives, impacting the overall performance. $\pi_{switch}$ detects the false positive by checking if $\Tilde{l}_t-\Tilde{q}_t < (1-q_{safe})$ and $\Tilde{l}+\Tilde{q}_t> 1.1$. If this condition is satisfied, $\pi_{switch}$ detects a false positive and speeds up the quadruped instead of slowing down.

We use as baseline policy \textbf{No Quantile} as in the simulation experiment. This policy modulates the robot speed only according to $\Tilde{l}_t$ and  $q_{safe}$, without any uncertainty associated with the prediction.

\subsection{Evaluation Results}
We evaluate the above policy on three closed trajectories with different difficulties. We use a triangle as the easy trajectory, a rectangle as the medium hard trajectory, and a hexagon as the hard trajectory. When the distance between the robot's estimated position and the last waypoint is lower than 0.1m and at least 75\% of waypoints are achieved, the robot stops tracking. As it's hard to obtain the ground-truth quadruped positions, we denote the tracking error as the distance between the start and end points of the quadruped and record the time to finish tracking. 

The predictor is trained with data collected with a fixed P gain of 1.5. Experiments \emph{in distribution} have $kp_{min}=1.2$, $kp_{max}=1.8$, therefore quite close to the predictor training data.  
Experiments \emph{out of distribution} have $kp_{min}=0.8$, $kp_{max}=4$, much beyond what the predictor was trained for. For both in-distribution and out-of-distribution experiments, we repeat each trajectory three times and compare the tracking error and time for our policy and the No Quantile baseline policy. 

The results are summarized in Figure~\ref{fig:slam_error},~\ref{fig:slam_time} and~\ref{fig:real_cov}. As shown in Figure~\ref{fig:slam_error}, in both in-distribution and out-of-distribution scenarios, the policy using both the prediction and the quantile achieves a lower tracking error with a slightly longer tracking time. For the easy trajectory, $\pi_{switch}$ achieves a similar tracking error to the baseline policy for in distribution experiment. The benefit of quantile shows up in the out-of-distribution experiments, where tracking error difference is more significant. Similarly, for the hard trajectory, the benefits of the quantile show up more in the out-of-distribution scenarios. For the medium trajectory, $\pi_{switch}$ is much better at tracking error compared to the baseline policy for in distribution case. We attribute this to the bad performance of the predictor on this rectangle trajectory. As shown in Figure~\ref{fig:slam_time}, for easy and medium trajectory, $\pi_{switch}$ takes longer to track as it slows down more to avoid SLAM errors. $\pi_{switch}$ is faster than the No Quantile policy for the hard trajectory with out-of-distribution states. This is because the predictor is overly conservative on this out-of-distribution trajectory and generates many false positives. 
Figure~\ref{fig:real_cov} shows the average coverage over a window of 4s for a hard trajectory for both in-distribution and out-of-distribution states using $\pi_{switch}$. In both cases, the coverage is higher than the desired threshold of $1-\alpha=0.8$, empirically validating the formal guarantee from the conformal quantiles.

% \begin{algorithm}
% \caption{Quadruped SLAM Tracking Policy}\label{alg:slam}
% \begin{algorithmic}
% \Require $\alpha \geq 0$, $\eta \geq 0, \gamma > 0, (X_0,...X_N), \epsilon \geq 0$
% \State $i \gets 0$, $t \gets 0$ , $\hat{y} \gets 0$,  $\hat{X}_t=(\hat{x}_t,\hat{y}_t)$,  $X_i=(x_i,y_i), O_t$

% \While{$||X_N - \hat{X}_t||\geq \epsilon$  or $i<0.75*N$ }
% $\hat{y} = \pi^p(O_i)$
% \If{$N$ is even}
%     \State $X \gets X \times X$
%     \State $N \gets \frac{N}{2}$  \Comment{This is a comment}
% \ElsIf{$N$ is odd}
%     \State $y \gets y \times X$
%     \State $N \gets N - 1$
% \EndIf
% \EndWhile
% \end{algorithmic}
% \end{algorithm}

\section{Conclusions}
We studied conformal policy learning, a method for safe robotic control under distribution shift.
The theoretical results prove that the conformal policy is sure to quickly exit regimes where its internal notion of uncertainty is flawed, causing its error rate to be too high.
Physical and simulation experiments validate these claims in robotic control setups and show practical benefits compared even to strong baselines under distribution shift.
Directions for future work include designing novel algorithms to improve the performance of RL controllers when provided with explicit uncertainty measures.
%%%%%%%%%%%%%%%%%%%%%%%%%%%%%%%%%%%%%%%%%%%%%%%%%%%%%%%%%%%%%%%%%%%%%%%%%%%%%%%%

%%%%%%%%%%%%%%%%%%%%%%%%%%%%%%%%%%%%%%%%%%%%%%%%%%%%%%%%%%%%%%%%%%%%%%%%%%%%%%%%

%%%%%%%%%%%%%%%%%%%%%%%%%%%%%%%%%%%%%%%%%%%%%%%%%%%%%%%%%%%%%%%%%%%%%%%%%%%%%%%%

% \balance
\bibliographystyle{IEEEtran}
\bibliography{references}% .bib

% \clearpage 
% \appendix
% \input{sections/appendix}
\end{document}

%% file: table.tex
\begin{table}
\vspace{10pt}
    \resizebox{\columnwidth}{!}{%
    \begin{tabular}{c|c|c|c|c|c|c|}
        \cline{2-7}
         & \multicolumn{2}{|c|}{Speed 20-30} & \multicolumn{2}{|c|}{Speed 20-40} & \multicolumn{2}{|c|}{Speed 20-50} \\
        \cline{2-7}
        \cline{2-7}
         & Duration & Speed & Duration & Speed &
         Duration & Speed\\
        \hline
        \multicolumn{1}{|c|}{$\pi_{speed}$} & 33.6 $\pm$ 11.0 & 29.0 $\pm$ 1.2 & 19.4 $\pm$ 11.7  & 35.1 $\pm$ 2.9 & 13.3 $\pm$ 9.0 & 36.0 $\pm$ 2.5 \\
        \hline

        \multicolumn{1}{|c|}{$\pi^{nq}_{RL}$} & 32.7 $\pm$ 11.5 & 29.3 $\pm$ 0.9 & 19.4 $\pm$ 11.3  & 34.5 $\pm$ 3.1 & 20.3 $\pm$ 11.7 & 34.1 $\pm$ 2.6 \\
        \hline
        
        \multicolumn{1}{|c|}{$\pi_{RL}$} & 32.2 $\pm$ 11.3 & 29.4 $\pm$ 0.8 & 25.0 $\pm$ 11.7  & 32.1 $\pm$ 3.0 & 20.1 $\pm$ 11.9 & 34.0 $\pm$ 2.1 \\
        \hline

        \multicolumn{1}{|c|}{$\pi^1_{RL}$} & 33.2 $\pm$ 11.3 & 29.3 $\pm$ 0.9 & 13.1 $\pm$ 8.6  & 36.0 $\pm$ 2.3 & 13.3 $\pm$ 9.0 & 36.0 $\pm$ 2.5 \\
        \hline
        
        \multicolumn{1}{|c|}{$\pi^2_{RL}$} & 34.4 $\pm$ 10.6 & 29.4 $\pm$ 0.9 & 16.5 $\pm$ 9.5  & 35.6 $\pm$ 2.8 & 16.7 $\pm$ 10.6 & 36.3 $\pm$ 2.6 \\
        \hline

        \multicolumn{1}{|c|}{$\pi^3_{RL}$} & 33.5 $\pm$ 11.6 & 28.8 $\pm$ 1.4 & 18.0 $\pm$ 10.8  & 33.8 $\pm$ 3.5 & 19.5 $\pm$ 11.8 & 34.2 $\pm$ 3.1 \\
        \hline

        \multicolumn{1}{|c|}{$\pi_{switch}$} & 38.7 $\pm$ 4.5 & 24.2 $\pm$ 1.5 & 37.8 $\pm$ 4.7  & 23.9 $\pm$ 1.8 &34.9 $\pm$ 8.0& 25.3$\pm$ 2.3 \\
        \hline
        
    \end{tabular}
    }
        \caption{Comparison between different variants of conformal policies trained end-to-end with RL to $\pi_{speed}$ and $\pi_{switch}$. Interestingly, no agents trained with RL manages to
use conformal quantiles as effectively as a switching policy.}
        \vspace{-15pt}
        \label{table:policy}
\end{table}

%% file: slam_fig.tex
% \begin{figure}
%      \centering
%      \begin{subfigure}[b]{0.235\textwidth}
%          \centering
%          \includegraphics[width=\textwidth]{figures/error_id.png}
         
%          \label{fig:error_id}
%      \end{subfigure}
%      \hfill
%      \begin{subfigure}[b]{0.235\textwidth}
%          \centering
%          \includegraphics[width=\textwidth]{figures/error_od.png}

%          \label{fig:error_od}
%      \end{subfigure}

%         \caption{Trajectory tracking error for P gain within (left) and out of training distribution. The}
%         \label{fig:slam_error}
% \end{figure}

% \begin{figure}
%     \centering
%     \includegraphics[width=0.49\textwidth]{figures/error.png}
%     \caption{Trajectory tracking error for P gain within (left) and out of training distribution.}
%     \label{fig:slam_error}
% \end{figure}
% \begin{figure}
%     \centering
%     \includegraphics[width=0.49\textwidth]{figures/time.png}
%     \caption{Trajectory tracking time for P gain within (left) and out of training distribution.}
%     \label{fig:slam_time}
% \end{figure}

\begin{figure*}[!tbp]
  \centering
  \begin{minipage}[b]{0.49\textwidth}
    \includegraphics[width=\textwidth]{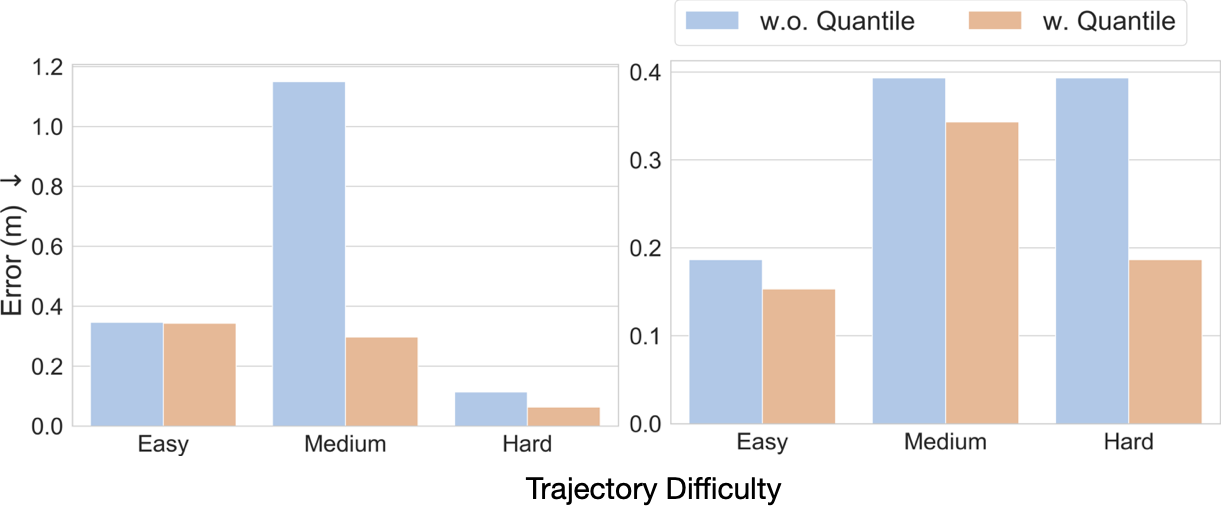}
     \caption{Tracking error for in (left) and out (right) distribution.}\label{fig:slam_error}
  \end{minipage}
  \hfill
  \begin{minipage}[b]{0.49\textwidth}
    \includegraphics[width=\textwidth]{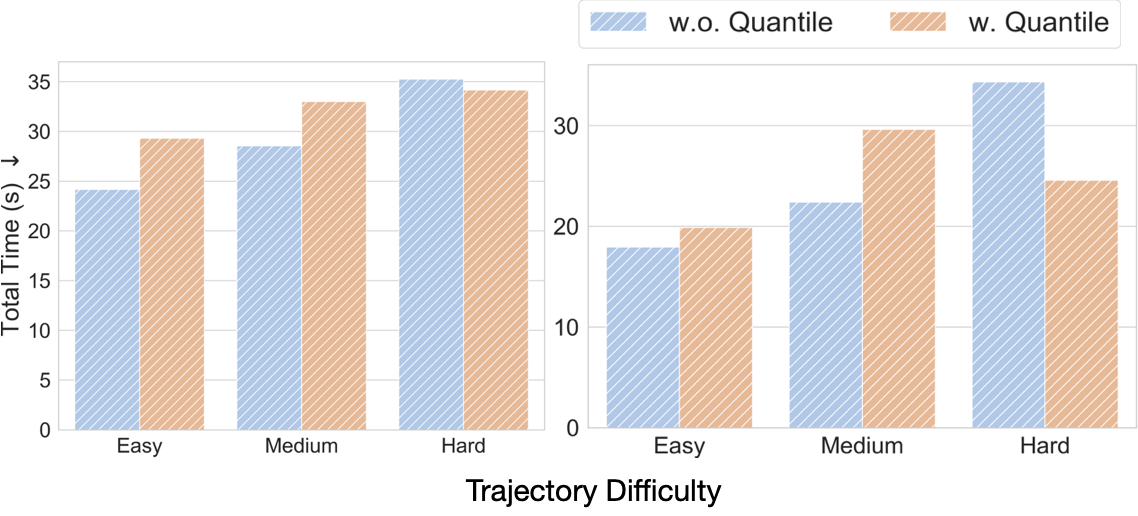}
     \caption{Tracking time for in (left) and out (right) distribution.}
    \label{fig:slam_time}
  \end{minipage}
  \vspace{-20pt}
\end{figure*}

\begin{figure}
    \centering
    \includegraphics[width=0.5\textwidth]{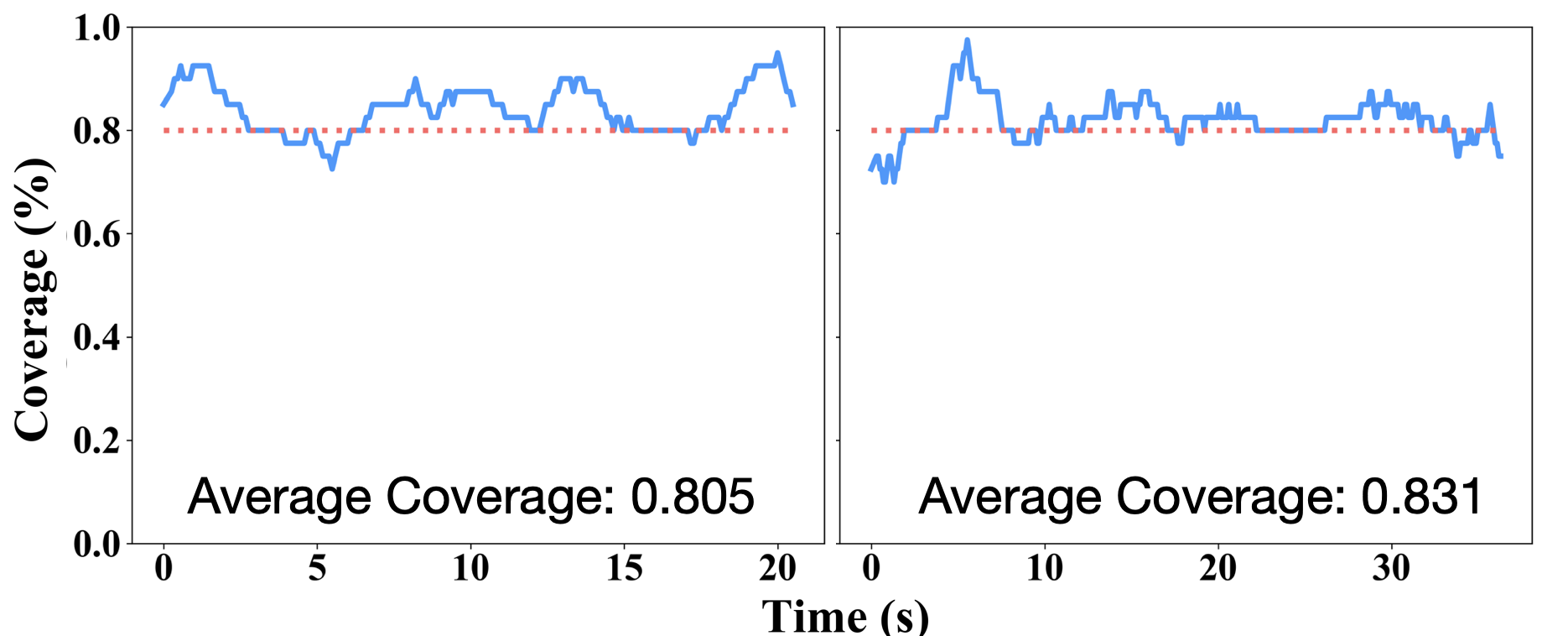}
    \caption{Plots of coverage with a sliding window size of 4 seconds generated with our method. On the left, we show the coverage for SLAM error prediction for a hard trajectory with within distribution P gains. On the right, we show that with out of the distribution P gains. In both cases, the average coverage over the entire rollout (labeled at the bottom right) is higher than the specified value 0.8, marked by the red dotted line. }
    \label{fig:real_cov}
    \vspace{-20pt}
\end{figure}